\pdfoutput=1

\documentclass[11pt]{article}

\usepackage{ACL2023}

\usepackage{amssymb}
\usepackage{times}
\usepackage{latexsym}
\usepackage{graphicx}
\usepackage{amsmath}
\usepackage{amsfonts}
\usepackage{multirow}
\usepackage{booktabs}
\usepackage{stmaryrd}

\usepackage[T1]{fontenc}

\usepackage[utf8]{inputenc}

\usepackage{microtype}

\usepackage{inconsolata}

\title{\emph{Don't Lose Yourself!}\\ Empathetic Response Generation via Explicit Self-Other Awareness}

\author{Weixiang Zhao, Yanyan Zhao\thanks{\ \ Corresponding author} \ , Xin Lu, Bing Qin \\
        Research Center for Social Computing and Information Retrieval\\
        Harbin Institute of Technology, China\\
        \texttt{\{wxzhao, yyzhao\}@ir.hit.edu.cn}}

\begin{document}
\maketitle
\begin{abstract}
As a critical step to achieve human-like chatbots, empathetic response generation has attained increasing interests. Previous attempts are incomplete and not sufficient enough to elicit empathy because they only stay on the initial stage of empathy to automatically sense and simulate the feelings and thoughts of others via other-awareness. However, they ignore to include self-awareness to consider the own views of the self in their responses, which is a crucial process to achieve the empathy. To this end, we propose to generate \textbf{Emp}athetic response with explicit \textbf{S}elf-\textbf{O}ther \textbf{A}wareness (\textbf{EmpSOA}). Specifically, three stages, self-other differentiation, self-other modulation and self-other generation, are devised to clearly maintain, regulate and inject the self-other aware information into the process of empathetic response generation. Both automatic and human evaluations on the benchmark dataset demonstrate the superiority of EmpSOA to generate more empathetic responses. Our source code is available at \url{https://github.com/circle-hit/EmpSOA}.
\end{abstract}

\section{Introduction}

Empathy is a desirable trait of the engaging human conversation and is considered as the key step to human-like chatbots. In this paper, we focus on the task of empathetic response generation \citep{ED_data}, which understands the feelings and situations of the user and responses properly.

According to one of the most influential theories of empathy proposed by \citet{empathy_definition}, \emph{empathy is the ability to sense the others' private world as if it were our own, \textbf{but without losing the ``as if'' condition.} } Previous works \citep{moel,empdg,cem} mainly focus on the prior part of the definition, which is referred as emotional contagion \citep{hatfield1993emotional}, and they automatically mimic the thoughts and feelings of the speaker to converge emotionally.

\begin{figure}
\centering
\includegraphics[width=1.0\columnwidth]{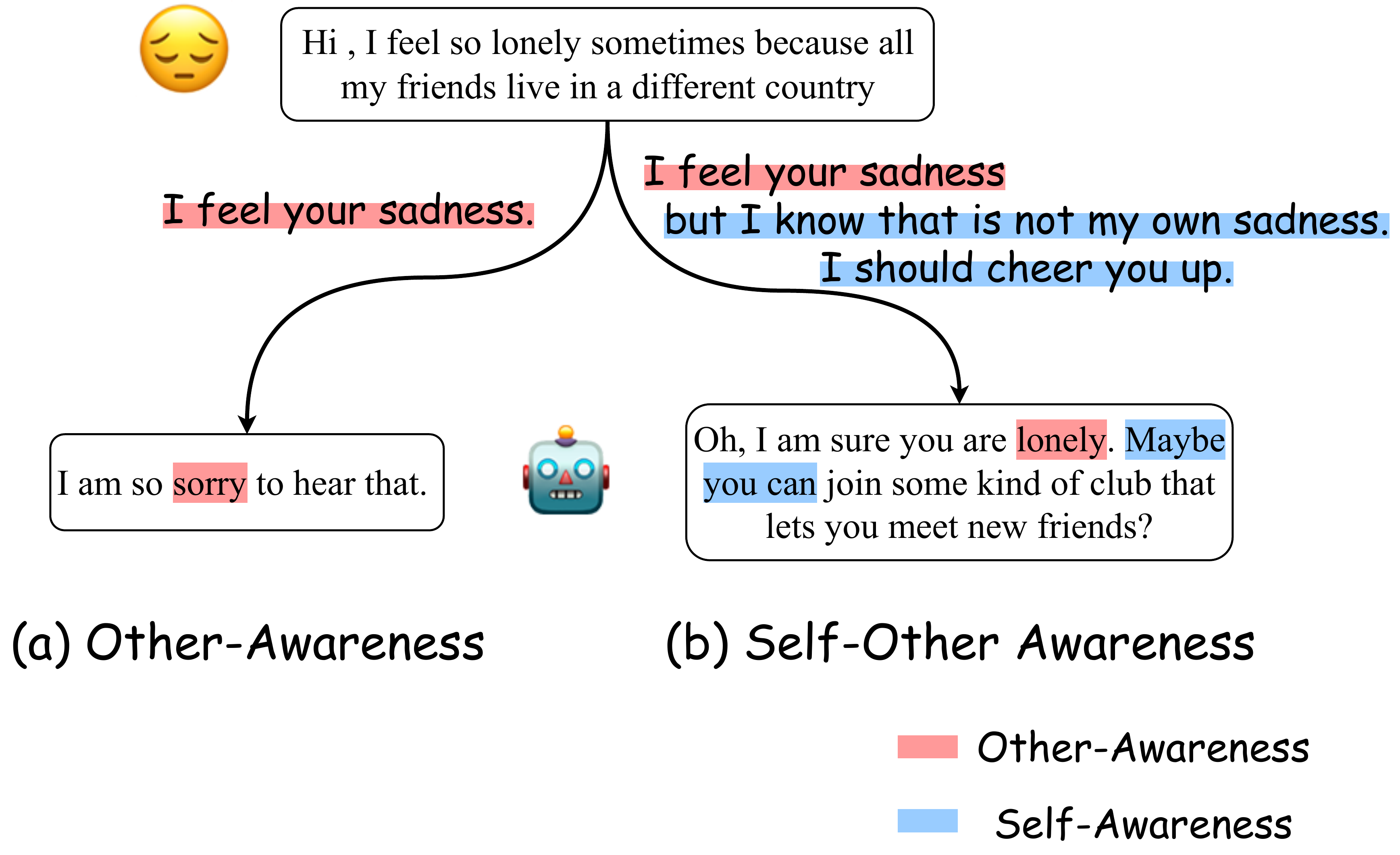}
\caption{An example for the self-other awareness during an empathetic conversation from the \textsc{EmpatheticDialogues} \citep{ED_data} dataset. The user and the system are referred as the other and the self.}
\label{example}
\end{figure}

However, emotional contagion is only the initial component that precedes empathy and what makes empathy distinct from emotion contagion is a critical process called \textbf{self-other awareness} \citep{decety2006human}. Thus, it is not sufficient enough for previous attempts to convey empathy because they only perform the single \textbf{other-awareness} to perceive the emotion and situation from the other and generate responses coupled with the same perceived emotion. As shown in the left part of Figure~\ref{example}, with other-awareness, the self succeed to \emph{feel the sadness} of the other. But the complete emotional overlap would induce the self to reinforce such sadness with the response of \emph{so sorry to hear that}, which is not the goal of empathy.

And it is of great necessity to take the own view of the self into consideration to maintain the ``as if'' condition with a clear \textbf{self-awareness}, which is a conscious process to maintain and modulate the own views of the self during the empathetic interaction. In the right part of Figure~\ref{example}, with the incorporation of the self-awareness, the self consciously holds in mind that \emph{this it not my own sadness} and I am responsible to \emph{cheer you up}. Thus, the explicit self-other awareness plays pivotal roles in disentangling feelings and views of the self and the other, which constitutes a crucial perspective of empathy and largely contributes to generate more empathetic responses, especially when the other is in negative emotional states.

To this end, we propose to generate \textbf{Emp}athetic response with explicit \textbf{S}elf-\textbf{O}ther \textbf{A}wareness (\textbf{EmpSOA}). Inspired by the conceptual framework of information flow involved in human empathy \citep{decety2006human}, we make such processes computable and abstract three stages in EmpSOA, named Self-Other Differentiation (SOD), Self-Other Modulation (SOM) and Self-Other Generation (SOG). Specifically, in SOD, we construct two heterogeneous graphs with four types of nodes to maintain the self-awareness representation and other-awareness representation, respectively. Among them, commonsense knowledge from COMET \citep{comet} is leveraged to manifest the fine-grained emotional and cognitive statuses of the self and the other. Further, we dynamically control the contributions of the self-other awareness representations in SOM and inject them into the process of empathetic response generation in SOG. Experimental results of both automatic and manual evaluations on the benchmark dataset demonstrate the superiority of EmpSOA to generate more empathetic responses.

The main contributions of this work are summarized as follows:
\begin{itemize}
    \item We propose to generate empathetic responses via explicit self-other awareness, which constitutes a critical perspective of empathy.
    \item We devise a novel model EmpSOA to clearly maintain, modulate and inject the self-other aware information into the process of empathetic response generation.
    \item Results of extensive experiments on the benchmark dataset demonstrate the effectiveness of EmpSOA to identify the exact emotion of the other and generate more empathetic response.
\end{itemize}

\section{Related Work}
\subsection{Empathetic Response Generation}
Endowing empathy to the dialogue systems has gained more and more attentions recently. For previous attempts on empathetic response generation, we divide them into two categories according to whether they incorporate both affection and cognition aspects of empathy. On the one hand, most existing works \citep{alam2018annotating,ED_data,moel,mime,empdg,kemp,wang2022empathetic} only consider the affective aspect of empathy to understand the emotional state of the other and converge emotionally. On the other hand, \citet{cem} propose to comprehensively understand the emotional feelings and cognitive situations of the other with commonsense knowledge incorporated.

However, all previous methods only perceive the emotional or cognitive states of the other by the single other-awareness, ignoring to explicitly incorporate self-awareness to make an appropriate empathetic response with own views of the self. 

\subsection{Emotional Dialogue Generation}
Emotion has been proven to be the key factor of achieving more engaging dialogue systems. Previous works explore two ways of incorporating emotion into dialogue generation. On the one hand, the generation-based methods \citep{ecm,emoji,CDL} are proposed to generate emotional responses given a specified emotion label. On the other hand, retrieval-based \citep{QiuSLSL0020,LuTZ021} methods aim to obtain emotional responses from candidates retrieved from the response repository. However, expressing the specified emotion in responses is merely the fundamental goal to achieve emotional dialogue systems, which is lack of the understanding for user's feelings and situations required by the empathetic response generation.

\begin{figure*}[htbp]
\centering
\includegraphics[width=0.8\textwidth]{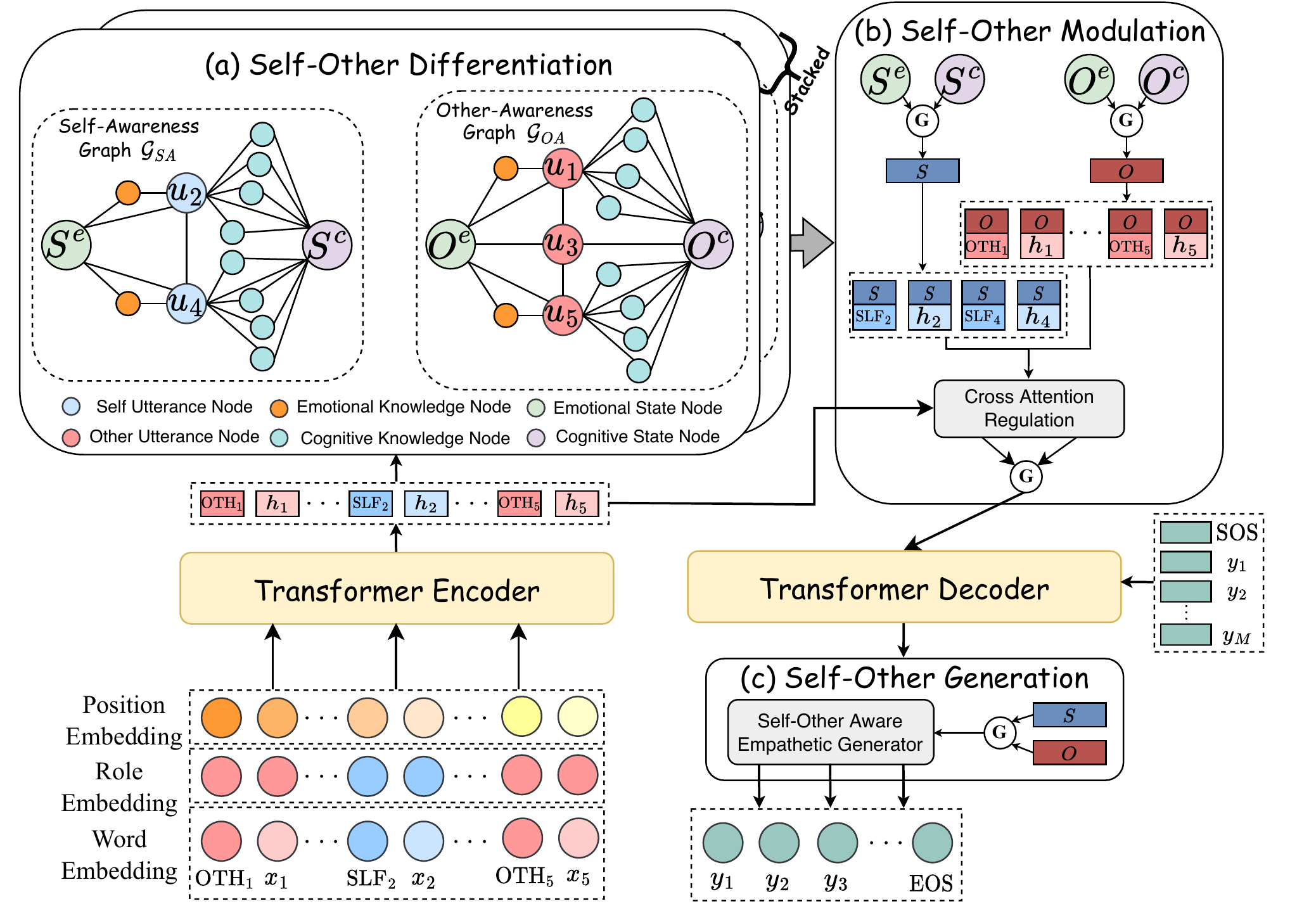}
\caption{The overall architecture of our proposed EmpSOA model, which mainly consists of three modules: (a) Self-Other Differentiation; (b) Self-Other Modulation and (c) Self-Other Generation.}
\label{model}
\end{figure*}

\section{Methodology}

\subsection{Task Definition}
First, we define the task of empathetic response generation. Formally, let $D=[X_1, X_2, \cdots, X_N]$ denotes a dialogue history with $N$ utterances between the user (the other) and the system (the self), where the $i$-th utterance $X_i = [w^i_1, w^i_2 \cdots, w^i_m]$ is a sequence of $m$ words. Besides, each conversation is provided with an emotion label $e$ from the total 32 available emotions to signal what the emotional tone that the other is grounded on. The goal is to generate the next utterance $Y$ from the stand of the self that is coherent to the dialogue history $D$ and empathetic to the other's situation and feeling.

\subsection{Overview of the Architecture}
We display the overall architecture of EmpSOA in Figure~\ref{model}. We abstract three main stages from the conceptual framework \footnote{More details about the conceptual framework of human empathy are provided in appendix file.} of information flow involved in human empathy \citep{decety2006human} and make them computabel in EmpSOA, including (a) Self-Other Differentiation (SOD), (b) Self-Other Modulation (SOM), and (c) Self-Other Generation (SOG). We first clearly disentangle the emotional and cognitive states of the self and the other to maintain the self- and other-awareness representations individually in SOD. Then in SOM, they are dynamically modulated and controlled to make different contributions to the self-other aware contextual information obtained from the context encoder. Finally, such self- and other-awareness representations are explicitly injected into the generation process in SOG to obtain the empathetic responses from views of both the self and the other.

\subsection{Self-Other Aware Context Encoder}
We adopt Transformer encoder \citep{trans} to obtain the contextual representations of the dialogue history. Following previous works \citep{kemp,cem}, the dialogue is flattened into a word sequence. To make the encoder aware of the self-other distinction in the encoding phase, we append two special tokens, $[\mathrm{SLF}]$ and $[\mathrm{OTH}]$, to the beginning of each utterance from the self and the other, respectively. Further, role embedding is added to supplement extra self-other aware information. The final input of the self-other aware context encoder are the sum of word embedding, role embedding and position embedding:
\begin{equation}
    H^{so} = \textrm{Encoder}(E_w + E_r + E_p)
\end{equation}
where $H^{so} \in \mathbb{R}^{N \times d_h}$ and $d_h$ is the hidden dimension of the self-other aware context encoder.

\subsection{Self-Other Differentiation}
As mentioned above, the clear self-other awareness constitutes a crucial perspective of genuine empathy. To achieve this, we first devise self-other differentiation (SOD). Specifically, we construct two heterogeneous graphs, named self-awareness graph $\mathcal{G}_{SA}$ and other-awareness graph $\mathcal{G}_{OA}$, to disentangle and maintain self- and other-awareness representations separately. Inspired by \citet{cem}, both awareness representations consist of emotional and cognitive aspects. And we leverage commonsense knowledge from the external knowledge base ATOMIC \citep{atomic} to imply the fine-grained emotional and cognitive knowledge of the self and the other at each dialogue turn. Such knowledge is highly related to the personal mental states and it has been widely used in many emotional dialogue-related tasks \citep{ghosal2020cosmic,zhao2022knowledge,cauain,misc,cem,peng}.

\paragraph{Graph Construction.} Since the way of constructing $\mathcal{G}_{SA}$ and $\mathcal{G}_{OA}$ is symmetrical, we only elaborate the formation of $\mathcal{G}_{SA}$ for simplicity.

\textbf{Nodes:} There are four types of heterogeneous nodes in $\mathcal{G}_{SA}$ to form the node set $\mathcal{V}_{SA}$, including (1) utterance nodes $u_i$, which are utterances in the dialogue history from the turn of the self; (2) external knowledge nodes $e_i$ and $c_i$, which are the commonsense knowledge to imply emotional feelings and cognitive situations of the self at the fine-grained level; (3) emotional state node $S^e$ and (4) cognitive state node $S^c$ of the self.

\textbf{Edges:} We build edges $\mathcal{E}_{SA}$ among these nodes to connect (1) adjacent utterance nodes; (2) each utterance node with its corresponding two external knowledge nodes; (3) the emotional state node with all utterance nodes and emotional knowledge nodes and (4) the cognitive state node with all utterance nodes and cognitive knowledge nodes.

\paragraph{Graph Initialization.}  We also take the self-awareness graph $\mathcal{G}_{SA}$ as an example to describe how to initialize the four types of nodes. And the initialization ways and types of commonsense knowledge are same in both $\mathcal{G}_{SA}$ and $\mathcal{G}_{OA}$.

For \textbf{utterance nodes} $u_i$, we obtain the corresponding hidden states of special tokens SLF$_i$ from the self-other aware contextual representation $H^{so}$.

For \textbf{external knowledge nodes}, the generative commonsense transformer model COMET \citep{comet} is adopted to obtain the emotional and cognitive knowledge. We select relation types \textit{xReact} to manifest the emotional feelings and $\left\{\textit{xIntent},\textit{xNeed},\textit{xWant},\textit{xEffect}\right\}$ \footnote{Please refer to the appendix file for the detailed definitions of all the relations.} to infer the cognitive situations at each dialogue turn of the self, which are consistent with those used in \citet{cem}. Specifically, we adopt the BART-based \citep{bart} variation of COMET, which is trained on the ATOMIC-2020 dataset \citep{atomic2020}. And given each utterance $X_i$ belonging to the self to form the input format $(X_i, r, [\textrm{GEN}])$, COMET would generate descriptions of inferential content under the relation $r$. Then hidden state representations from the last layer of COMET are obtained to initialize $c_i$ and $e_i$.

For \textbf{emotional state node} and \textbf{cognitive state node} of the self, we randomly initialize them.

\paragraph{Self-Other Aware Graph Attention.}
Based on the self-awareness graph $\mathcal{G}_{SA}$ and the other-awareness graph $\mathcal{G}_{OA}$, we apply the multi-head graph attention mechanism to update the node representations in each graph. Concretely, the graph attention operated on the node representation to update it from the information of other neighbourhoods can be written as:
\begin{equation}
    \hat{v_i} =\bigparallel_{n=1}^H(\sum_{j \in N_i} \alpha_{ij} W_v^n v_j)
\end{equation}
where $\bigparallel$ denotes the concatenation of $H$ attention heads, $N_i$ is the neighbors of node $i$, and $W_v^n$ is the linear transformation.

The attention weight of $n$-th head $\alpha^n_{ij}$ is utilised to measure the importance and relevance between the current node and its neighbours:
\begin{equation}
    \alpha^n_{ij} = \frac{\mathrm{exp}((W_q^n v_i)^\top (W_k^n v_j))}{\sum_{j^{\prime} \in N_i} \mathrm{exp}((W_q^n v_i)^\top (W_k^n v_{j^{\prime}})}
\end{equation}
where $W_q^n$ and $W_k^n$ are both linear transformation.

After $L$ stacked layers of multi-head self-other aware graph attention, both emotional state node and cognitive state node of the self and the other would aggregate the fine-grained self-other aware information, achieving the clear self-other differentiation for the following parts of our model.

\paragraph{Emotion Perception.} Since we are provided with the golden emotion label of each conversation, an emotion classifier is devised to accurately comprehend the emotional state of the other. Unlike previous attempts that perform emotion classification without a clear differentiation between the emotional state of the self and the other \citep{empdg,kemp,cem}, we exactly focus on the emotional state of the other $O^e$ and the average of corresponding $[\mathrm{OTH}]$ tokens from $H^{so}$:
\begin{equation}
    h_e = \textrm{Average}(OTHs) + O^e
\end{equation}
where $OTHs$ is the sequence of $\mathrm{OTH}_i$ derived from $H^{so}$. Then, we pass ${h}_e$ through a linear layer followed by the softmax operation to generate the emotion category distribution $P_{emo}$:

\begin{equation}
    P_{emo} = \textrm{softmax}(W^eh_e)
\end{equation}
where $P_{emo} \in \mathbb{R}^{n_e}$, $W^e \in \mathbb{R}^{d_h \times n_e}$ and $n_e$ is the number of total available emotion categories.

During training, we perform the parameter learning by minimizing the Cross-Entropy (CE) loss between the emotion category distribution $P_{emo}$ and the ground truth label $e$:
\begin{equation}
    \mathcal{L}_{emo} = - \log(P_{emo}(e))
\end{equation}

\subsection{Self-Other Modulation}
Through SOD, we differentiate and maintain the self- and other-awareness representations. And what followed is Self-Other Modulation (SOM) module, a conscious and controlled process to determine to what extent we pay attention to them.

First, the emotional state and the cognitive state are dynamically fused by a gate mechanism to obtain the joint self-awareness representation $S$:
\begin{equation}
\begin{aligned}
    S &= g^s \odot S^e + (1-g^s) \odot S^c \\
    g^s &= \sigma([S^e; S^c]W^s + b^s) \\
    \end{aligned}
\end{equation}

Similarly, the fused other-awareness representation $O$ can be obtained by:
\begin{equation}
\begin{aligned}
    O &= g^o \odot O^e + (1-g^o) \odot O^c \\
    g^o &= \sigma([O^e; O^c]W^o + b^o) \\
    \end{aligned}
\end{equation}
where $\odot$ is the element-wise multiplication, $\sigma$ is the sigmoid activation function and $W^s \in \mathbb{R}^{2d_h \times d_h}$, $W^o \in \mathbb{R}^{2d_h \times d_h}$, $b^s \in \mathbb{R}^{d_h}$ and $b^o \in \mathbb{R}^{d_h}$ are all trainable parameters.

Then, to refine the context with the self-other aware information, we respectively concatenate $S$ and $O$ to their corresponding self-other aware contextual representation $H^{so}$ at the token level:
\begin{equation}
    \hat{H^s}[i] = S \oplus H^{s}[i]
\end{equation}
\begin{equation}
    \hat{H^o}[i] = O \oplus H^{o}[i]
\end{equation}
where $H^s$ and $H^o$ are the slices of $H^{so}$ and belongs to the self and the other, respectively.

And Feed Forward Neural Network \citep{trans} is applied to perform the self-other aware context refinement in the point-wise way:
\begin{equation}
    \tilde{H^s} = \textrm{FFN}^s(\hat{H^s})
\end{equation}
\begin{equation}
    \tilde{H^o} = \textrm{FFN}^o(\hat{H^o})
\end{equation}

Finally, we adopt the cross attention mechanism to control and modulate the contribution of self-awareness context and other-awareness context:
\begin{equation}
    C^s = \textrm{CROSS-ATT}^s(H^{so}, \tilde{H^s})
\end{equation}
\begin{equation}
    C^o = \textrm{CROSS-ATT}^o(H^{so}, \tilde{H^o})
\end{equation}

And a gate is applied to obtain the modulated self-other aware contextual representation:
\begin{equation}
\begin{aligned}
    C^{so} &= g \odot C^s + (1-g) \odot C^o \\
    g &= \sigma([C^s; C^o]W^m + b^m) \\
    \end{aligned}
\end{equation}
where $W^m \in \mathbb{R}^{2d_h \times d_h}$ and $b^m \in \mathbb{R}^{d_h}$ are trainable parameters.

\subsection{Self-Other Generation}
Finally, we devise the self-other generation (SOG) to inject the self-other aware information into the process of empathetic response generation.

For the target response $Y=[y_1, y_2, \cdots, y_M]$, to generate the $t$-th word $y_t$, we firstly feed the previous generated words $y_{1:t-1}$ into the vanilla Transformer decoder \citep{trans}. It is worth to mention that the input of cross attention is modified to the self-other aware contextual representation $C^{so}$ derived from the SOM module:
\begin{equation}
    h_t = \textrm{Decoder}(E_{y < t}, C^{so})
\end{equation}
where $E_{y < t}$ denotes the embeddings of the generated words before the time step $t$.

Then to make the generation process grounded on both views of the self and the other, we dynamically inject self- and other- awareness representations via the fusion of them and the hidden representation $h_t$ of the $t$-th token:
\begin{equation}
\begin{aligned}
    h &= h_t + g^f \odot S + (1-g^f) \odot O \\
    g^y &= \sigma([h_t; S; O]W^f + b^f) \\
    \end{aligned}
\end{equation}
where $W^f \in \mathbb{R}^{3d_h \times d_h}$ and $b^f \in \mathbb{R}^{d_h}$ are trainable parameters.

The distribution over the vocabulary for the $t$-th token can be obtained by a softmax layer:
\begin{equation}
    P(y_t \mid y_{<t}, D) = \textrm{softmax}(Wh + b)
\end{equation}
where $D$ is the input dialogue history, $W \in \mathbb{R}^{\lvert V \rvert \times d_h}$ and $V$ is the vocabulary size.

We utilise the standard negative log-likelihood as the generation loss function:
\begin{equation}
L_{gen}=-\sum_{t=1}^{M} \log P\left(y_{t} \mid D, y_{<t}\right).
\end{equation}

A multi-task learning framework is adopted to jointly minimize the emotion perception loss, the generation loss and the diversity loss proposed by \citet{cem}:
\begin{equation}
L=\gamma_{1} L_{emo}+\gamma_{2} L_{gen}+\gamma_{3} L_{div}
\end{equation}
where $\gamma_{1}$, $\gamma_{2}$ and $\gamma_{3}$ are hyper-parameters.

\section{Experiments}

\subsection{Dataset}
We conduct our experiments on \textsc{EmpatheticDialogues} dataset \citep{ED_data}. It is a large-scale multi-turn dataset with 25k empathetic conversations collected on the Amazon Mechanical Turk. In this dataset, empathetic conversations are carried out between a speaker and a listener (which is referred as the other and the self in this paper). In addition, 32 evenly distributed emotion labels are provided to signal the personal emotional feelings of the other. We use the same dataset split of 8:1:1 train/valid/test with that in \citet{ED_data}.

\subsection{Baselines}
We compare our proposed EmpSOA with the following competitive baselines.
\begin{itemize}
    \item \textbf{Transformer} \cite{trans}: The vanilla Transformer-based encoder-decoder generation model.
    \item \textbf{Multi-Task Transformer (Multi-TRS)} \cite{ED_data}: A variation of the vanilla Transformer with an additional structure to perform emotion perception.
    \item \textbf{MoEL} \cite{moel}: A Transformer-based model that captures emotions of the other and outputs an emotion distribution with multi decoders. By softly combining the output emotion distribution, each decoder is optimized to react to certain emotions, and generate an empathetic response.
    \item \textbf{MIME} \cite{mime}: Another Transformer-based model that mimics the emotion of the other to a varying degree by grouping emotions into two clusters. Stochasticity is introduced to yield emotionally more varied empathetic responses.
    \item \textbf{EmpDG} \cite{empdg}: An adversarial empathetic response generation model that exploits both the coarse-grained dialogue-level and fine-grained token-level emotions, and the interactive user feedback.
    \item \textbf{KEMP} \cite{kemp}: An encoder-decoder model that leverages external knowledge, including commonsense knowledge and emotional lexical knowledge, to explicitly understand and express emotions in empathetic dialogue generation.
    \item \textbf{CEM} \cite{cem}: For the first time to focus on both affection and cognition of empathy and leverage commonsense to draw more information about the user’s situation. It uses this additional information to
    enhance the empathy expression in generated responses.
\end{itemize}

\subsection{Implementation Details}
We implemented all the baselines and our model with 5 random runs. 300-dimensional pre-trained GloVE vectors \citep{glove} are adopted to initialize the word embeddings and shared between the encoder and the decoder. The hidden dimension $d_h$ is also set to 300 and the number of attention heads in SOD graph attention and SOM cross attention are 6. Loss weights $\gamma_{1}$, $\gamma_{2}$ and $\gamma_{3}$ are set to 1, 1, and 1.5, respectively. Adam \cite{adam} optimizer with $\beta_1=0.9$ and $\beta_2=0.98$ is used for training. Following \citet{trans}, we vary the learning rate during the training process with the initial learning rate of 0.0001. Early stopping is applied when training. And the training process is performed on one single Tesla V100 GPU with a mini-batch size of 16. For inference, we use a batch size of 1 and a maximum of 30 decoding steps for all models.

\subsection{Evaluation Metrics}

\paragraph{Automatic Evaluation.} We apply three kinds of automatic metrics for evaluation: (1) Perplexity (\textbf{PPL}) measures the general quality of the generated responses; (2) Distinct-$n$ (\textbf{Dist}-$n$) \citep{dist} evaluates the diversity of the generated responses by measuring the ratio of unique $n$-grams; (3) Accuracy (\textbf{Acc}) of the emotion perception is utilised to evaluate the model capability for understanding the emotional state of the other. Following CEM, we do not report the word overlap-based automatic metrics such as BLEU \citep{bleu} because they are not appropriate for evaluating dialogue systems \citep{LiuLSNCP16}.

\paragraph{Human Evaluation.} Following \citet{cem}, we conduct the aspect-based pairwise preference test for human evaluation. Specifically, 100 response pairs generated by EmpSOA and baselines are randomly sampled. Then we ask 5 professional annotators to choose the better response following three aspects: (1) Coherence (\textbf{Coh.}): which response is more coherent and relevant to the dialogue history; (2) Empathy (\textbf{Emp.}): which response is more empathetic to show a better understanding of the other's feelings and situations; (3) Informativeness (\textbf{Inf.}): which response contains more information related to the dialogue history.

\begin{table}
\footnotesize
\centering
\begin{tabular}{ccccc}
\toprule
\textbf{Model} & \textbf{PPL} & \textbf{Acc} & \textbf{Dist-1} & \textbf{Dist-2} \\
\midrule
Transformer &37.62 &- &0.43 &1.98\\
Multi-TRS   &37.73 &32.86 &0.43 &1.92\\
MoEL &36.73 &31.28 &0.56 &2.82\\
MIME &37.37 &29.86 &0.40 &1.66\\
EmpDG &37.38 &30.79 &0.42 &1.87\\
KEMP &36.39 &36.57 &0.61 &2.65\\
CEM &36.49 &37.34 &0.60 &2.85\\
\midrule
EmpSOA (Ours) &\textbf{35.02} &\textbf{48.32} &\textbf{0.71} &\textbf{3.96} \\
\bottomrule& 
\end{tabular}
\caption{Comparison of our model against state-of-the-art baselines in terms of the automatic evaluation. The best results among all models are highlighted in bold.}
\label{tab2}
\end{table}

\section{Results and Analysis}

\subsection{Overall Results}
\paragraph{Automatic Evaluation.} Illustrated in Table~\ref{tab2}, EmpSOA achieves the new state-of-the-art automatic evaluation results. Benefiting from the clear self-other awareness to maintain, modulate and inject the self- and other-awareness representations, EmpSOA is capable of generating empathetic responses of higher quality with the lowest PPL compared to all the baselines. In addition, the improvement on Dist-1 and Dist-2 indicates the superiority of EmpSOA in terms of generating more informative responses at the unigrams and bigrams level. Finally, although the similar external emotional knowledge is explored in CEM and KEMP, the prominent performance on emotion perception of the other can be ascribed to the explicit disentanglement of the emotional states between the self and the other.

\paragraph{Human Evaluation.} As shown in Table~\ref{tab3}, EmpSOA significantly outperforms three competitive baselines in terms of all three aspects, which demonstrates the superiority of EmpSOA to generate more empathetic and informative responses with explicit self-other awareness. In addition, we adopt the Fleiss's kappa \citep{fleiss1971measuring} to measure the overall inter-rater agreement. And the agreement ratio falls in the range of [0.41, 0.6], which denotes the moderate agreement.

\begin{table}
\footnotesize
  \centering
    \begin{tabular}{c c c c}
        \toprule
        \textbf{Comparisons} &
        \textbf{Aspects} &
        \textbf{Win} &
        \textbf{Lose} \\
        \midrule
        &  Coh. &\textbf{47.0$^\ddag$} &38.0\\
        EmpSOA vs. CEM& Emp. &\textbf{47.0$^\ddag$} &37.4\\
        & Inf. &\textbf{47.6$^\ddag$} &36.6\\
        \midrule
        &  Coh. &\textbf{45.6$^\ddag$} &34.8\\
        EmpSOA vs. KEMP& Emp. &\textbf{45.8$^\ddag$} &33.8\\
        & Inf. &\textbf{46.4$^\ddag$} &34.6\\
        \midrule
         &  Coh. &\textbf{48.2$^\ddag$} &33.4\\
        EmpSOA vs. EmpDG& Emp. &\textbf{49.8$^\ddag$} &35.6\\
        & Inf. &\textbf{47.8$^\ddag$} &33.2\\
        \bottomrule
    \end{tabular}
  \caption{
    Human evaluation results (\%). Ties are not shown. ${\ddag}$ represent significant improvement with $p$-value $<0.05$.
  }
  \label{tab3}
\end{table}

\begin{table}
\footnotesize
\centering
\begin{tabular}{lcccc}
\toprule
\textbf{Model} & \textbf{PPL} & \textbf{Acc} & \textbf{Dist-1} & \textbf{Dist-2}\\
\midrule
EmpSOA &\textbf{35.02} &\textbf{48.32} &\textbf{0.71} &\textbf{3.96} \\
-SOG & 36.65 &46.36 &0.57 &2.59\\
-SOG \& SOM &37.43 &45.59 &0.57 &2.51\\
-SOG \& SOM \& SOD &37.18 &34.71 &0.52 &2.19\\
\bottomrule
\end{tabular}
\caption{Results of ablation study. SOG, SOM, SOD refer to the self-other generation, self-other modulation and self-other differentiation module, respectively.}
\label{tab4}
\end{table}

\begin{table*}
\small
\centering
\begin{tabular}{c|l}
\toprule
\textbf{Emotion} & Lonely \\
\textbf{Context} & Hi, I feel so lonely sometimes because all my friends live in a different country.\\
\midrule
\textbf{MIME} & I am sorry to hear that. Do you have any friends? \\
\textbf{EmpDG} & That is sad. I am sorry to hear that.\\
\textbf{KEMP} & I am so sorry to hear that. \\
\textbf{CEM} & I am so sorry to hear that. \\
\midrule
\textbf{EmpSOA} & I am sorry to hear that. Maybe you should try to talk to someone in the city and meet some new people.\\
\midrule
\textbf{Ground-Truth} & Oh, I am sure you are lonely. Maybe you can join some kind of club that lets you meet new friends?\\
\bottomrule
\end{tabular}

\caption{Case study of the generated empathetic responses by our proposed EmpSOA and the baselines.}
\label{tab6}
\end{table*}

\begin{table}
\footnotesize
\centering
\begin{tabular}{ccccc}
\toprule
\textbf{Model} & \textbf{PPL} & \textbf{Acc} & \textbf{Dist-1} & \textbf{Dist-2}\\
\midrule
EmpSOA &\textbf{35.02} &\textbf{48.32} &\textbf{0.71} &\textbf{3.96} \\
EmpNA &35.88 &47.44 &0.69 &3.34\\
EmpOA &35.78 &47.10 &0.63 &3.18\\
EmpSA &38.44 &35.07 &0.51 &2.03\\
\bottomrule
\end{tabular}
\caption{Results of deeper analysis on self-other awareness with three variants of EmpSOA.}
\label{tab5}
\end{table}

\subsection{Ablation Study}
We conduct ablation studies to verify the effectiveness of the three key modules, SOD, SOM and SOG, proposed in EmpSOA to achieve self-other awareness. Since they are highly correlated with each other, we remove each one of them according to the order of SOG, SOM and SOD individually.

\paragraph{Effect of Self-Other Generation.} To investigate the impact of SOG module in generating self-other aware empathetic response, we discard the dynamic fusion of self- and other-awareness representations in each decoding step. Results are displayed in the second row in Table~\ref{tab4}. Without the explicit injection of self-other aware information, the general quality and diversity of generation drops significantly. It manifests that it is of vital importance to explicitly offer the model the self-other aware information to generate more empathetic generation from the perspectives of both the self and the other.

\paragraph{Effect of Self-Other Modulation.} Subsequently, we remove the SOM module to study the effectiveness of the modulation for the self-other aware information. The dropped results shown in the third row of Table~\ref{tab4} prove the importance to dynamically control and modulate different contributions of self- and other-awareness representations. Further, it reminds us that it is not sufficient enough to generate empathetic response just by the clear self-other differentiation without any self-other aware information incorporating into the generation process.

\paragraph{Effect of Self-Other Differentiation.} Finally, the SOD module is discarded and results are shown in the last row. The significant decrease of emotion perception accuracy indicates that SOD make remarkable contribution to disentangle and perceive the emotion of the other. In addition, it is worth to mention that there is no any self-other aware related module in the current model. And compared to the complete EmpSOA, all results of the automatic evaluation decrease significantly, which supports our motivation that the clear self-other awareness contributes the crucial aspect of empathy.

\subsection{Deeper Analysis on Self-Other Awareness}
In this section, we demonstrate the in-depth analysis on how the explicit self-other awareness leads to more empathetic responses. Results are shown in Table~\ref{tab5}. Concretely, three variants of EmpSOA are implemented, including \textbf{Emp}athetic response generation with \textbf{N}o \textbf{A}wareness (EmpNA), with \textbf{O}ther \textbf{A}wareness (EmpOA) and with \textbf{S}elf \textbf{A}wareness (EmpSA). We will elaborate each one of them.

For \textbf{EmpNA}, we merge the self- and other-awareness graphs to construct a single heterogeneous graph without the explicit differentiation between the emotional and cognitive state of the self and the other in the SOD module. Thus, the original self-awareness $S$ and the other-awareness $O$ are replaced by a joint representation and it is fed into the following SOM and SOG. Through this, empathetic response would be generated without any self-other aware information. The decreased performance on generation quality and diversity confirms our constructed motivation to perform empathetic responses with self-other awareness.

For \textbf{EmpOA} and \textbf{EmpSA}, although we still differentiate the self-other awareness in the SOD module, only one of the self-awareness $S$ or the other-awareness $O$ is applied to SOM and SOG. Automatic evaluation results of both EmpOA and EmpSA decrease to a certain degree, which further verifies our claim to consider self-other awareness simultaneously when generating empathetic responses. Interestingly, the performance of EmpSA is much worsen than the other three models, which indicates that being selfish and over-focused on ourselves would neglect the feelings of the other, resulting in an improper way to elicit empathy.

\subsection{Case Study}
In Table~\ref{tab6}, we show a case with responses generated by EmpSOA and the four baselines.
It can be observed that all the models succeed to perceive the exact emotional state of the other and express \textit{sorry} to achieve the emotional consensus. However, it is not sufficient enough to elicit empathy only in this way. Through the explicit self-other awareness, EmpSOA not only accurately reaches the emotional state of the other via other-awareness, but also attempts to stay conscious to avoid being overwhelmed by the feelings of the other and provide the valuable suggestions on \textit{meeting new friends} via self-awareness, which is highly consistent with the empathy expressed in the ground-truth.

\section{Conclusion and Future Work}

In this paper, we propose EmpSOA to generate empathetic responses via explicit self-other awareness. Three stages including Self-Other Differentiation (SOD), Self-Other Modulation (SOM) and Self-Other Generation (SOG) are devised to achieve this goal. Experimental results on both automatic and human evaluation demonstrate the superiority of EmpSOA to generate more empathetic responses.

In the future, we will explore the theory of self-other awareness in tasks that specified to elicit the positive emotion of the other.

\section{Limitations}
There are three points to discuss and they may inspire further investigation. First, since the length of empathetic conversations in the current benchmark dataset \textsc{EmpatheticDialogues} \citep{ED_data} is relatively short, the theory of self-other awareness could be explored under the circumstance of long conversations to maintain the self-awareness of chatbots for the long run. Second, for the better comprehension of self-other awareness, it is helpful to introduce more commonsense knowledge of higher quality. Finally, current automatic evaluation metrics are still not rational and proper to measure the ability of empathy. It is desirable to build better evaluation metrics for empathetic responses.

\section{Ethics Statement}
The open-source benchmark dataset \textsc{EmpatheticDialogues} \citep{ED_data} used in our experiments is collected by employed crowd-sourced workers, with user privacy protected and no personal information involved. Besides, the participants in our human evaluation are volunteered transparently informed of our research intent, with reasonable wages paid.

\section*{Acknowledgements}
We thank the anonymous reviewers for their insightful comments and suggestions. This work was supported by the National Key RD Program of China via grant  2021YFF0901602 and the National Natural Science Foundation of China (NSFC) via grant 62176078.

\bibliography{anthology,custom}

\begin{thebibliography}{32}
\expandafter\ifx\csname natexlab\endcsname\relax\def\natexlab#1{#1}\fi

\bibitem[{Alam et~al.(2018)Alam, Danieli, and Riccardi}]{alam2018annotating}
Firoj Alam, Morena Danieli, and Giuseppe Riccardi. 2018.
\newblock Annotating and modeling empathy in spoken conversations.
\newblock \emph{Computer Speech \& Language}, 50:40--61.

\bibitem[{Bosselut et~al.(2019)Bosselut, Rashkin, Sap, Malaviya, Celikyilmaz,
  and Choi}]{comet}
Antoine Bosselut, Hannah Rashkin, Maarten Sap, Chaitanya Malaviya, Asli
  Celikyilmaz, and Yejin Choi. 2019.
\newblock \href {https://doi.org/10.18653/v1/p19-1470} {{COMET:} commonsense
  transformers for automatic knowledge graph construction}.
\newblock In \emph{Proceedings of the 57th Conference of the Association for
  Computational Linguistics, {ACL} 2019, Florence, Italy, July 28- August 2,
  2019, Volume 1: Long Papers}, pages 4762--4779. Association for Computational
  Linguistics.

\bibitem[{Decety and Lamm(2006)}]{decety2006human}
Jean Decety and Claus Lamm. 2006.
\newblock Human empathy through the lens of social neuroscience.
\newblock \emph{TheScientificWorldJOURNAL}, 6:1146--1163.

\bibitem[{Fleiss(1971)}]{fleiss1971measuring}
Joseph~L Fleiss. 1971.
\newblock Measuring nominal scale agreement among many raters.
\newblock \emph{Psychological bulletin}, 76(5):378.

\bibitem[{Ghosal et~al.(2020)Ghosal, Majumder, Gelbukh, Mihalcea, and
  Poria}]{ghosal2020cosmic}
Deepanway Ghosal, Navonil Majumder, Alexander Gelbukh, Rada Mihalcea, and
  Soujanya Poria. 2020.
\newblock Cosmic: Commonsense knowledge for emotion identification in
  conversations.
\newblock \emph{arXiv preprint arXiv:2010.02795}.

\bibitem[{Hatfield et~al.(1993)Hatfield, Cacioppo, and
  Rapson}]{hatfield1993emotional}
Elaine Hatfield, John~T Cacioppo, and Richard~L Rapson. 1993.
\newblock Emotional contagion.
\newblock \emph{Current directions in psychological science}, 2(3):96--100.

\bibitem[{Hwang et~al.(2021)Hwang, Bhagavatula, Bras, Da, Sakaguchi, Bosselut,
  and Choi}]{atomic2020}
Jena~D. Hwang, Chandra Bhagavatula, Ronan~Le Bras, Jeff Da, Keisuke Sakaguchi,
  Antoine Bosselut, and Yejin Choi. 2021.
\newblock \href {https://ojs.aaai.org/index.php/AAAI/article/view/16792}
  {(comet-) atomic 2020: On symbolic and neural commonsense knowledge graphs}.
\newblock In \emph{Thirty-Fifth {AAAI} Conference on Artificial Intelligence,
  {AAAI} 2021, Thirty-Third Conference on Innovative Applications of Artificial
  Intelligence, {IAAI} 2021, The Eleventh Symposium on Educational Advances in
  Artificial Intelligence, {EAAI} 2021, Virtual Event, February 2-9, 2021},
  pages 6384--6392. {AAAI} Press.

\bibitem[{Kingma and Ba(2015)}]{adam}
Diederik~P. Kingma and Jimmy Ba. 2015.
\newblock \href {http://arxiv.org/abs/1412.6980} {Adam: {A} method for
  stochastic optimization}.
\newblock In \emph{3rd International Conference on Learning Representations,
  {ICLR} 2015, San Diego, CA, USA, May 7-9, 2015, Conference Track
  Proceedings}.

\bibitem[{Lewis et~al.(2020)Lewis, Liu, Goyal, Ghazvininejad, Mohamed, Levy,
  Stoyanov, and Zettlemoyer}]{bart}
Mike Lewis, Yinhan Liu, Naman Goyal, Marjan Ghazvininejad, Abdelrahman Mohamed,
  Omer Levy, Veselin Stoyanov, and Luke Zettlemoyer. 2020.
\newblock {BART:} denoising sequence-to-sequence pre-training for natural
  language generation, translation, and comprehension.
\newblock In \emph{Proceedings of the 58th Annual Meeting of the Association
  for Computational Linguistics, {ACL} 2020, Online, July 5-10, 2020}, pages
  7871--7880. Association for Computational Linguistics.

\bibitem[{Li et~al.(2016)Li, Galley, Brockett, Gao, and Dolan}]{dist}
Jiwei Li, Michel Galley, Chris Brockett, Jianfeng Gao, and Bill Dolan. 2016.
\newblock \href {https://doi.org/10.18653/v1/n16-1014} {A diversity-promoting
  objective function for neural conversation models}.
\newblock In \emph{{NAACL} {HLT} 2016, The 2016 Conference of the North
  American Chapter of the Association for Computational Linguistics: Human
  Language Technologies, San Diego California, USA, June 12-17, 2016}, pages
  110--119. The Association for Computational Linguistics.

\bibitem[{Li et~al.(2020)Li, Chen, Ren, Ren, Tu, and Chen}]{empdg}
Qintong Li, Hongshen Chen, Zhaochun Ren, Pengjie Ren, Zhaopeng Tu, and Zhumin
  Chen. 2020.
\newblock \href {https://doi.org/10.18653/v1/2020.coling-main.394} {Empdg:
  Multi-resolution interactive empathetic dialogue generation}.
\newblock In \emph{Proceedings of the 28th International Conference on
  Computational Linguistics, {COLING} 2020, Barcelona, Spain (Online), December
  8-13, 2020}, pages 4454--4466. International Committee on Computational
  Linguistics.

\bibitem[{Li et~al.(2022)Li, Li, Ren, Ren, and Chen}]{kemp}
Qintong Li, Piji Li, Zhaochun Ren, Pengjie Ren, and Zhumin Chen. 2022.
\newblock \href {https://ojs.aaai.org/index.php/AAAI/article/view/21347}
  {Knowledge bridging for empathetic dialogue generation}.
\newblock In \emph{Thirty-Sixth {AAAI} Conference on Artificial Intelligence,
  {AAAI} 2022, Thirty-Fourth Conference on Innovative Applications of
  Artificial Intelligence, {IAAI} 2022, The Twelveth Symposium on Educational
  Advances in Artificial Intelligence, {EAAI} 2022 Virtual Event, February 22 -
  March 1, 2022}, pages 10993--11001. {AAAI} Press.

\bibitem[{Lin et~al.(2019)Lin, Madotto, Shin, Xu, and Fung}]{moel}
Zhaojiang Lin, Andrea Madotto, Jamin Shin, Peng Xu, and Pascale Fung. 2019.
\newblock \href {https://doi.org/10.18653/v1/D19-1012} {Moel: Mixture of
  empathetic listeners}.
\newblock In \emph{Proceedings of the 2019 Conference on Empirical Methods in
  Natural Language Processing and the 9th International Joint Conference on
  Natural Language Processing, {EMNLP-IJCNLP} 2019, Hong Kong, China, November
  3-7, 2019}, pages 121--132. Association for Computational Linguistics.

\bibitem[{Liu et~al.(2016)Liu, Lowe, Serban, Noseworthy, Charlin, and
  Pineau}]{LiuLSNCP16}
Chia{-}Wei Liu, Ryan Lowe, Iulian Serban, Michael Noseworthy, Laurent Charlin,
  and Joelle Pineau. 2016.
\newblock \href {https://doi.org/10.18653/v1/d16-1230} {How {NOT} to evaluate
  your dialogue system: An empirical study of unsupervised evaluation metrics
  for dialogue response generation}.
\newblock In \emph{Proceedings of the 2016 Conference on Empirical Methods in
  Natural Language Processing, {EMNLP} 2016, Austin, Texas, USA, November 1-4,
  2016}, pages 2122--2132. The Association for Computational Linguistics.

\bibitem[{Lu et~al.(2021)Lu, Tian, Zhao, and Qin}]{LuTZ021}
Xin Lu, Yijian Tian, Yanyan Zhao, and Bing Qin. 2021.
\newblock \href {https://doi.org/10.18653/v1/2021.findings-emnlp.168}
  {Retrieve, discriminate and rewrite: {A} simple and effective framework for
  obtaining affective response in retrieval-based chatbots}.
\newblock In \emph{Findings of the Association for Computational Linguistics:
  {EMNLP} 2021, Virtual Event / Punta Cana, Dominican Republic, 16-20 November,
  2021}, pages 1956--1969. Association for Computational Linguistics.

\bibitem[{Majumder et~al.(2020)Majumder, Hong, Peng, Lu, Ghosal, Gelbukh,
  Mihalcea, and Poria}]{mime}
Navonil Majumder, Pengfei Hong, Shanshan Peng, Jiankun Lu, Deepanway Ghosal,
  Alexander~F. Gelbukh, Rada Mihalcea, and Soujanya Poria. 2020.
\newblock \href {https://doi.org/10.18653/v1/2020.emnlp-main.721} {{MIME:}
  mimicking emotions for empathetic response generation}.
\newblock In \emph{Proceedings of the 2020 Conference on Empirical Methods in
  Natural Language Processing, {EMNLP} 2020, Online, November 16-20, 2020},
  pages 8968--8979. Association for Computational Linguistics.

\bibitem[{Papineni et~al.(2002)Papineni, Roukos, Ward, and Zhu}]{bleu}
Kishore Papineni, Salim Roukos, Todd Ward, and Wei{-}Jing Zhu. 2002.
\newblock \href {https://doi.org/10.3115/1073083.1073135} {Bleu: a method for
  automatic evaluation of machine translation}.
\newblock In \emph{Proceedings of the 40th Annual Meeting of the Association
  for Computational Linguistics, July 6-12, 2002, Philadelphia, PA, {USA}},
  pages 311--318. {ACL}.

\bibitem[{Peng et~al.(2022)Peng, Hu, Xing, Xie, Sun, and Li}]{peng}
Wei Peng, Yue Hu, Luxi Xing, Yuqiang Xie, Yajing Sun, and Yunpeng Li. 2022.
\newblock \href {https://doi.org/10.24963/ijcai.2022/600} {Control globally,
  understand locally: {A} global-to-local hierarchical graph network for
  emotional support conversation}.
\newblock In \emph{Proceedings of the Thirty-First International Joint
  Conference on Artificial Intelligence, {IJCAI} 2022, Vienna, Austria, 23-29
  July 2022}, pages 4324--4330. ijcai.org.

\bibitem[{Pennington et~al.(2014)Pennington, Socher, and Manning}]{glove}
Jeffrey Pennington, Richard Socher, and Christopher~D. Manning. 2014.
\newblock \href {https://doi.org/10.3115/v1/d14-1162} {Glove: Global vectors
  for word representation}.
\newblock In \emph{Proceedings of the 2014 Conference on Empirical Methods in
  Natural Language Processing, {EMNLP} 2014, October 25-29, 2014, Doha, Qatar,
  {A} meeting of SIGDAT, a Special Interest Group of the {ACL}}, pages
  1532--1543. {ACL}.

\bibitem[{Qiu et~al.(2020)Qiu, Shiu, Lin, Song, Liu, Zhao, and
  Yan}]{QiuSLSL0020}
Lisong Qiu, Yingwai Shiu, Pingping Lin, Ruihua Song, Yue Liu, Dongyan Zhao, and
  Rui Yan. 2020.
\newblock \href {https://doi.org/10.1145/3397271.3401108} {What if bots feel
  moods?}
\newblock In \emph{Proceedings of the 43rd International {ACM} {SIGIR}
  conference on research and development in Information Retrieval, {SIGIR}
  2020, Virtual Event, China, July 25-30, 2020}, pages 1161--1170. {ACM}.

\bibitem[{Rashkin et~al.(2019)Rashkin, Smith, Li, and Boureau}]{ED_data}
Hannah Rashkin, Eric~Michael Smith, Margaret Li, and Y{-}Lan Boureau. 2019.
\newblock \href {https://doi.org/10.18653/v1/p19-1534} {Towards empathetic
  open-domain conversation models: {A} new benchmark and dataset}.
\newblock In \emph{Proceedings of the 57th Conference of the Association for
  Computational Linguistics, {ACL} 2019, Florence, Italy, July 28- August 2,
  2019, Volume 1: Long Papers}, pages 5370--5381. Association for Computational
  Linguistics.

\bibitem[{Rogers(1992)}]{empathy_definition}
Carl~R Rogers. 1992.
\newblock The necessary and sufficient conditions of therapeutic personality
  change.
\newblock \emph{Journal of consulting and clinical psychology}, 60(6):827.

\bibitem[{Sabour et~al.(2022)Sabour, Zheng, and Huang}]{cem}
Sahand Sabour, Chujie Zheng, and Minlie Huang. 2022.
\newblock \href {https://ojs.aaai.org/index.php/AAAI/article/view/21373}
  {{CEM:} commonsense-aware empathetic response generation}.
\newblock In \emph{Thirty-Sixth {AAAI} Conference on Artificial Intelligence,
  {AAAI} 2022, Thirty-Fourth Conference on Innovative Applications of
  Artificial Intelligence, {IAAI} 2022, The Twelveth Symposium on Educational
  Advances in Artificial Intelligence, {EAAI} 2022 Virtual Event, February 22 -
  March 1, 2022}, pages 11229--11237. {AAAI} Press.

\bibitem[{Sap et~al.(2019)Sap, Bras, Allaway, Bhagavatula, Lourie, Rashkin,
  Roof, Smith, and Choi}]{atomic}
Maarten Sap, Ronan~Le Bras, Emily Allaway, Chandra Bhagavatula, Nicholas
  Lourie, Hannah Rashkin, Brendan Roof, Noah~A. Smith, and Yejin Choi. 2019.
\newblock \href {https://doi.org/10.1609/aaai.v33i01.33013027} {{ATOMIC:} an
  atlas of machine commonsense for if-then reasoning}.
\newblock In \emph{The Thirty-Third {AAAI} Conference on Artificial
  Intelligence, {AAAI} 2019, The Thirty-First Innovative Applications of
  Artificial Intelligence Conference, {IAAI} 2019, The Ninth {AAAI} Symposium
  on Educational Advances in Artificial Intelligence, {EAAI} 2019, Honolulu,
  Hawaii, USA, January 27 - February 1, 2019}, pages 3027--3035. {AAAI} Press.

\bibitem[{Shen and Feng(2020)}]{CDL}
Lei Shen and Yang Feng. 2020.
\newblock \href {https://doi.org/10.18653/v1/2020.acl-main.52} {{CDL:}
  curriculum dual learning for emotion-controllable response generation}.
\newblock In \emph{Proceedings of the 58th Annual Meeting of the Association
  for Computational Linguistics, {ACL} 2020, Online, July 5-10, 2020}, pages
  556--566. Association for Computational Linguistics.

\bibitem[{Tu et~al.(2022)Tu, Li, Cui, Wang, Wen, and Yan}]{misc}
Quan Tu, Yanran Li, Jianwei Cui, Bin Wang, Ji-Rong Wen, and Rui Yan. 2022.
\newblock \href {https://doi.org/10.18653/v1/2022.acl-long.25} {{MISC}: A mixed
  strategy-aware model integrating {COMET} for emotional support conversation}.
\newblock In \emph{Proceedings of the 60th Annual Meeting of the Association
  for Computational Linguistics (Volume 1: Long Papers)}, pages 308--319,
  Dublin, Ireland. Association for Computational Linguistics.

\bibitem[{Vaswani et~al.(2017)Vaswani, Shazeer, Parmar, Uszkoreit, Jones,
  Gomez, Kaiser, and Polosukhin}]{trans}
Ashish Vaswani, Noam Shazeer, Niki Parmar, Jakob Uszkoreit, Llion Jones,
  Aidan~N. Gomez, Lukasz Kaiser, and Illia Polosukhin. 2017.
\newblock \href
  {https://proceedings.neurips.cc/paper/2017/hash/3f5ee243547dee91fbd053c1c4a845aa-Abstract.html}
  {Attention is all you need}.
\newblock In \emph{Advances in Neural Information Processing Systems 30: Annual
  Conference on Neural Information Processing Systems 2017, December 4-9, 2017,
  Long Beach, CA, {USA}}, pages 5998--6008.

\bibitem[{Wang et~al.(2022)Wang, Li, Lin, Meng, Yang, Wang, and
  Zhou}]{wang2022empathetic}
Lanrui Wang, Jiangnan Li, Zheng Lin, Fandong Meng, Chenxu Yang, Weiping Wang,
  and Jie Zhou. 2022.
\newblock Empathetic dialogue generation via sensitive emotion recognition and
  sensible knowledge selection.
\newblock \emph{arXiv preprint arXiv:2210.11715}.

\bibitem[{Zhao et~al.(2022{\natexlab{a}})Zhao, Zhao, Li, and
  Qin}]{zhao2022knowledge}
Weixiang Zhao, Yanyan Zhao, Zhuojun Li, and Bing Qin. 2022{\natexlab{a}}.
\newblock Knowledge-bridged causal interaction network for causal emotion
  entailment.
\newblock \emph{arXiv preprint arXiv:2212.02995}.

\bibitem[{Zhao et~al.(2022{\natexlab{b}})Zhao, Zhao, and Lu}]{cauain}
Weixiang Zhao, Yanyan Zhao, and Xin Lu. 2022{\natexlab{b}}.
\newblock \href {https://doi.org/10.24963/ijcai.2022/628} {Cauain: Causal aware
  interaction network for emotion recognition in conversations}.
\newblock In \emph{Proceedings of the Thirty-First International Joint
  Conference on Artificial Intelligence, {IJCAI} 2022, Vienna, Austria, 23-29
  July 2022}, pages 4524--4530. ijcai.org.

\bibitem[{Zhou et~al.(2018)Zhou, Huang, Zhang, Zhu, and Liu}]{ecm}
Hao Zhou, Minlie Huang, Tianyang Zhang, Xiaoyan Zhu, and Bing Liu. 2018.
\newblock \href
  {https://www.aaai.org/ocs/index.php/AAAI/AAAI18/paper/view/16455} {Emotional
  chatting machine: Emotional conversation generation with internal and
  external memory}.
\newblock In \emph{Proceedings of the Thirty-Second {AAAI} Conference on
  Artificial Intelligence, (AAAI-18), the 30th innovative Applications of
  Artificial Intelligence (IAAI-18), and the 8th {AAAI} Symposium on
  Educational Advances in Artificial Intelligence (EAAI-18), New Orleans,
  Louisiana, USA, February 2-7, 2018}, pages 730--739. {AAAI} Press.

\bibitem[{Zhou and Wang(2018)}]{emoji}
Xianda Zhou and William~Yang Wang. 2018.
\newblock \href {https://doi.org/10.18653/v1/P18-1104} {Mojitalk: Generating
  emotional responses at scale}.
\newblock In \emph{Proceedings of the 56th Annual Meeting of the Association
  for Computational Linguistics, {ACL} 2018, Melbourne, Australia, July 15-20,
  2018, Volume 1: Long Papers}, pages 1128--1137. Association for Computational
  Linguistics.

\end{thebibliography}
\bibliographystyle{acl_natbib}

\newpage

\appendix

\section{Conceptual Framework of Information Flow in Human Empathy}
\begin{figure*}
\centering
\includegraphics[width=0.8\textwidth]{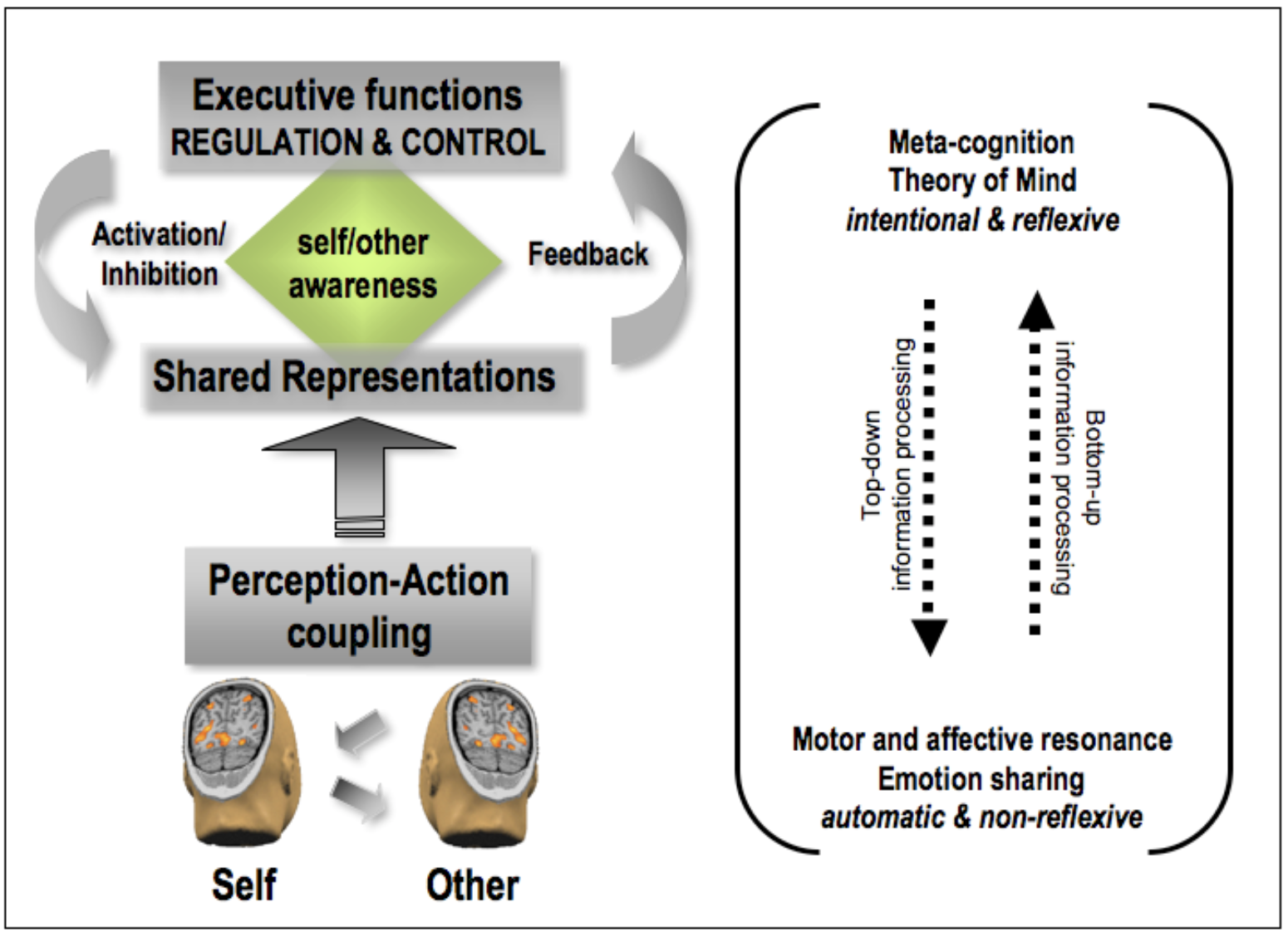}
\caption{Schematic representation of the bottom-up (i.e., direct matching between perception and action) and top-down (i.e., regulation and control) information processes involved in human empathy. These two levels of processing are interrelated. Top-down regulation, through executive functions, modulates low levels and adds flexibility, making the individual less dependent on external cues. The meta-cognitive feedback plays a crucial role in taking into account one’s own mental competence in order to react (or not) to the affective states of others. \citep{decety2006human}}
\label{human}
\end{figure*}
This framework \citep{decety2006human} shown in Figure \ref{human} considers empathy as a construct that accounts for a sense of similarity in the feelings experienced by self and other (such translations go both ways, from other-to-self and from self-to-other), without confusion between the two agents. It involves both bottom-up and top-down information processing. Furthermore, it combines representational aspects, i.e., memories that are localized in distributed neural networks that encode information and, when temporarily activated, enable access to this stored information, as well as processes, i.e., computational procedures that are localized and are independent of the nature or modality of the stimulus that is being processed.

Inspired by this framework, we regard the above bottom-up information flow, which performs a sense of similarity in the feelings experienced by self and other, as the Self-Other Differentiation (SOD) to maintain clear self-other awareness in EmpSOA. Moreover, the top-down regulation process is abstracted as the Self-Other Modulation (SOM) to control the weighted contributions of self-other awareness. Finally, the Self-Other Generation (SOG) is similar to the meta-cognitive feedback, taking into account the self- and other-awareness information to elicit empathetic responses.

\section{Description of ATOMIC Relations}
\label{sec:appendix}

ATOMIC \citep{atomic} is an atlas of everyday commonsense reasoning and organized through textual descriptions of inferential knowledge, where nine if-then relation types are proposed to distinguish causes vs. effects, agents vs. themes, voluntary vs. involuntary events, and actions vs. mental states. We give the brief definition of each relation.

\textbf{xIntent} Why does PersonX cause the event?

\textbf{xNeed} What does PersonX need to do before the event?

\textbf{xAttr} How would PersonX be described?

\textbf{xEffect} What effects does the event have on PersonX?

\textbf{xWant} What would PersonX likely want to do after the event?

\textbf{xReact} How does PersonX feel after the event?

\textbf{oReact} How does others' feel after the event?

\textbf{oWant} What would others likely want to do after the event?

\textbf{oEffect} What effects does the event have on others?

\end{document}